%% file: main.tex
\documentclass[conference]{IEEEtran}

\usepackage{amsmath,amssymb}
\usepackage{graphicx}
\usepackage{booktabs}
\usepackage{cite}
\usepackage{url}
\usepackage{placeins}
\title{Fault-Geometry-Aware Control Allocation for Emergency Quadrotor Landing under Single-Motor Loss of Effectiveness}

\author{
\small
\begin{tabular}{ccc}

\begin{tabular}{c}
\textbf{1\textsuperscript{st} Suhas S P}\\
\textit{Department of CSE}\\
Siddaganga Institute of Technology\\
Visvesvaraya Technological University\\
Tumakuru, Karnataka, India\\
1si22cs180@sit.ac.in
\end{tabular}

&

\begin{tabular}{c}
\textbf{2\textsuperscript{nd} Samridhi Salwan}\\
\textit{Department of Economics}\\
St. Joseph's University\\
Bengaluru, Karnataka, India\\
samridhisalwan7@gmail.com \\
\small ORCID: \url{https://orcid.org/0009-0003-8576-1916}
\end{tabular}

&

\begin{tabular}{c}
\textbf{3\textsuperscript{rd} Arshan Ali Khan}\\
\textit{Department of CSE}\\
Siddaganga Institute of Technology\\
Visvesvaraya Technological University\\
Tumakuru, Karnataka, India\\
1si22ci007@sit.ac.in
\end{tabular}

\\[3.2em]

\begin{tabular}{c}
\textbf{4\textsuperscript{th} Ujwal Pai Jakarbet}\\
\textit{Department of CSE}\\
Siddaganga Institute of Technology\\
Visvesvaraya Technological University\\
Tumakuru, Karnataka, India\\
1si22cs190@sit.ac.in
\end{tabular}

&

\begin{tabular}{c}
\textbf{5\textsuperscript{th} Savithramma R M}\\
\textit{Department of CSE}\\
Siddaganga Institute of Technology\\
Visvesvaraya Technological University\\
Tumakuru, Karnataka, India\\
savirmrl@sit.ac.in
\end{tabular}

&

\begin{tabular}{c}
\textbf{6\textsuperscript{th} Ashwini B P}\\
\textit{Department of CSE}\\
Siddaganga Institute of Technology\\
Visvesvaraya Technological University\\
Tumakuru, Karnataka, India\\
ashvinibp@sit.ac.in
\end{tabular}

\end{tabular}
}

\begin{document}
\maketitle

\begin{abstract}
Partial single-motor loss of effectiveness (LoE) changes the remaining control-authority geometry of a quadrotor and can change which control allocator performs best during emergency landing. We compare three allocators in Crazyflie software-in-the-loop simulation: bounded weighted least squares (reported as PINV), a finite-candidate residual allocator (QP-lite), and the same residual allocator with scoring parameters tuned offline by the Cross-Entropy Method (CEM). For this study, fault geometry is represented by the known failed-motor index. At $\eta=0.496$ remaining effectiveness, matched development trials select CEM for M1, M3, and M4 and PINV for M2; the M3 selection is a one-trial margin (29/30 versus 28/30). After freezing this routing map, the policy achieves 117/120 safe first-contact touchdowns on fresh seeds and 115/120 in a separate randomized-order integrated-supervisor evaluation. For the critical M2/PINV branch, Firth regression gives a nominal $\eta_{50}\approx0.4956$ with a within-model 95\% bootstrap interval of approximately $[0.4955,0.4956]$. One-factor-at-a-time $\pm10\%$ plant perturbations shift $\eta_{50}$ by up to 0.05495 for thrust coefficient and 0.04974 for mass; the most frequent violated first-contact criterion also changes from vertical speed in many conditions to angular rate for mass $-10\%$ and thrust-coefficient $+10\%$. At fixed $\eta=0.496$, PINV has a higher M2 safe count than frozen CEM in 9/11 tested plant states and ties in 2/11; the two ties are both 0/30. The experiments therefore distinguish motor-conditioned allocator selection from sensitivity to plant-model mismatch.
\end{abstract}

\begin{IEEEkeywords}
fault-tolerant control, quadrotor, control allocation, Crazyflie, actuator loss of effectiveness, emergency landing
\end{IEEEkeywords}

\input{sections/01_introduction}
\input{sections/02_related_work}
\input{sections/03_problem_formulation}
\input{sections/04_method}
\input{sections/05_experiments}
\input{sections/06_results}
\input{sections/07_discussion}
\input{sections/08_conclusion}

\FloatBarrier
\IEEEtriggeratref{19}
\bibliographystyle{IEEEtran}
\bibliography{references}
\end{document}

%% file: sections/01_introduction.tex
\section{Introduction}
\label{sec:introduction}

Quadrotors have little actuator redundancy: degradation of one motor changes the set of control effects available to the vehicle and can compromise landing safety. Rotor faults have therefore motivated fault-tolerant control (FTC), reduced-attitude recovery, control reconfiguration, and constrained control allocation \cite{saied2024review,mueller2014loss,lu2015active,merheb2017emergency,nan2022nmpc}. Control allocation is attractive because it separates nominal motion control from the redistribution of actuator commands under saturation or failure \cite{johansen2013survey,bodson2002allocation,harkegard2002active}.

This work asks a narrower question than general fault recovery: under the same partial single-motor LoE severity, does the best emergency allocator depend on which motor fails? We compare bounded WLS/PINV, finite-candidate QP-lite residual allocation, and an offline CEM-tuned QP-lite variant in CrazySim \cite{llanes2024crazysim}. In the implemented supervisor, ``fault geometry'' is represented only by the known failed-motor index; no online geometry estimator or fault detector is evaluated.

A second question is whether a controller-selection conclusion remains meaningful when the simulated plant is perturbed. We therefore separate three stages: development-set allocator comparison, frozen-policy validation on new seeds, and an M2-focused plant-sensitivity study. The contributions are: (1) matched evidence that allocator ranking depends on failed-motor location, with a close M3 comparison explicitly retained; (2) frozen motor-conditioned routing evaluated on fresh seeds and in an integrated randomized-order supervisor; and (3) nominal-boundary, OFAT, failure-criterion, and paired-routing analyses that quantify how M2/PINV behavior changes under the tested plant perturbations.

%% file: sections/02_related_work.tex
\section{Related Work}
\label{sec:related}

\subsection{Fault-Tolerant Quadrotor Control}
Complete rotor loss has been studied through reduced-attitude or free-yaw recovery \cite{mueller2014loss,ke2023uniform}, active fault detection/reconfiguration \cite{lu2015active}, emergency control allocation \cite{merheb2017emergency}, upset recovery \cite{sun2020upset}, and nonlinear MPC \cite{nan2022nmpc}. Partial actuator faults and model uncertainty have also motivated adaptive, sliding-mode, observer-based, and allocation-based FTC \cite{baldini2020actuator,zhu2019backstepping,wang2023adaptive}. These studies establish that multirotors can retain useful control authority after actuator faults. Our experiments instead hold the fault class fixed and compare multiple allocators under matched failed-motor locations before freezing a location-conditioned routing rule.

\subsection{Control Allocation and Simulation}
Control-allocation literature includes pseudoinverse methods, constrained weighted least squares, active-set methods, and secondary-objective formulations \cite{johansen2013survey,bodson2002allocation,harkegard2002active,harkegard2005redundancy}. Experimental and constrained-allocation studies address actuator saturation and bounds \cite{kirchengast2018allocation}, while multirotor-specific formulations explicitly incorporate rotor constraints \cite{santos2022allocation}. Gazebo \cite{koenig2004gazebo}, RotorS \cite{furrer2016rotors}, AirSim \cite{shah2018airsim}, and CrazySim \cite{llanes2024crazysim} provide simulation environments with different levels of vehicle and firmware integration. We use CrazySim because the final experiments execute Crazyflie firmware against Gazebo physics, allowing the same fault injection and bounded allocator implementation to be exercised in SITL. Firth bias-reduced logistic regression is used for boundary data with near-separation \cite{firth1993bias,heinze2002separation}; CEM is used only as an offline score-parameter tuner \cite{rubinstein2004cem}.

%% file: sections/03_problem_formulation.tex
\section{System and Problem Formulation}
\label{sec:problem}

\subsection{Fault and Safety Model}
Let $\mathbf{u}=[u_1,u_2,u_3,u_4]^\top$ be the nominal motor-command vector. For the single faulted motor $m$, the SITL LoE model applies $u_m^{\mathrm{eff}}=\eta u_m$; all $j\neq m$ retain $u_j^{\mathrm{eff}}=u_j$. Here $\eta=1$ is healthy and lower $\eta$ is more severe. The main controller-comparison and supervisor experiments use $\eta=0.496$.

\begin{figure*}[!t]
\centering
\includegraphics[width=0.94\textwidth]{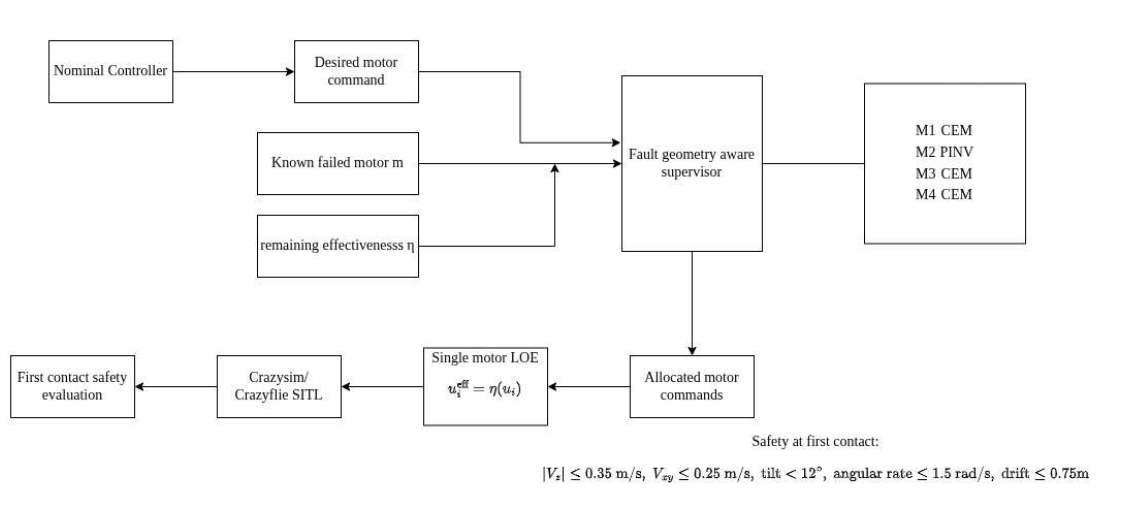}
\caption{Implemented emergency-landing architecture. Nominal motor commands and supplied fault information $(m,\eta)$ enter the frozen supervisor before the simulated single-motor LoE block.}
\label{fig:system}
\end{figure*}

The supervisor is supplied with the failed-motor identity $m$ and prescribed $\eta$; online fault detection/isolation is not implemented or evaluated. For the single-motor LoE class studied here, the motor index is used as a discrete label for the corresponding remaining actuator geometry. The resulting architecture is shown in Fig.~\ref{fig:system}.

A trial is evaluated at the first post-fault sample with $z\le0.03$~m. It is labeled safe only if ground contact occurs and all first-contact limits are satisfied: $|v_z|\le0.35$~m/s, $v_{xy}\le0.25$~m/s, $\theta_{rp}\le12^\circ$, $\|\boldsymbol{\omega}\|\le1.5$~rad/s, and $d_{xy}\le0.75$~m. Because one unsafe trial can violate several limits, criterion-violation counts are nonexclusive.

\subsection{Objectives}
We select a frozen routing map $g:m\mapsto\pi_m$, where $\pi_m\in\{\pi_{\mathrm{PINV}},\pi_{\mathrm{QP}},\pi_{\mathrm{CEM}}\}$, using development trials only. We then evaluate that frozen map on new seeds. Finally, for the M2/PINV branch, we estimate the nominal safe-touchdown transition, measure its OFAT sensitivity to five plant parameters, summarize which first-contact criteria are violated, and test whether the PINV--CEM ordering reverses at fixed $\eta=0.496$.

%% file: sections/04_method.tex
\section{Fault-Geometry-Aware Emergency Allocation}
\label{sec:allocation}

\subsection{Bounded WLS/PINV}
The bounded allocator acts in the Crazyflie PWM-command domain. With nominal motor command $\mathbf{u}_0$, define the legacy command-effectiveness matrix
\begin{equation}
A=\begin{bmatrix}1&1&1&1\\-1&-1&1&1\\1&-1&-1&1\\1&-1&1&-1\end{bmatrix},
\end{equation}
and let $D_m(\eta)$ be identity with diagonal entry $m$ replaced by $\eta$. The allocator solves
\begin{equation}
\mathbf{u}^*=\arg\min_{0\le\mathbf{u}\le65535}\!\|W(AD_m\mathbf{u}-A\mathbf{u}_0)\|_2^2+\lambda\|\mathbf{u}-\mathbf{u}_0\|_2^2,
\label{eq:wls_objective}
\end{equation}
with $W=\mathrm{diag}(1,1,1,0.2)$ and $\lambda=10^{-6}$. This is a PWM-domain command-effectiveness objective, not a physical force/wrench allocation model. The firmware enumerates all $3^4=81$ lower/free/upper active sets, rejects infeasible sets, and selects the feasible minimum. The experimental label ``PINV'' is retained for continuity with the project nomenclature, but the executed controller is bounded WLS rather than unconstrained inversion followed by clipping.

\subsection{QP-Lite and Offline CEM Tuning}
QP-lite applies $\mathbf{u}=\mathbf{u}_0+\mathbf{r}$ with $r_m=0$ and chooses from a fixed motor-specific residual library: seven candidates for M1/M4 and eight for M2/M3. At fault activation it evaluates local empirical surrogates of touchdown vertical speed, horizontal displacement, and tilt. In compact form,
\begin{align}
\hat V_z&=0.350+6[0.500-\eta]_+ +0.16[[-v_z]_+-0.05]_+ \nonumber\\
&\quad+0.06[0.35-z]_+-2.20\!\times\!10^{-6}r_{opp}\nonumber\\
&\quad-1.50\!\times\!10^{-7}r_{sup}+p_m(r_{opp}),\label{eq:qplite_vz}\\
\hat d_{xy}&=\sqrt{x^2+y^2}+0.60\sqrt{v_x^2+v_y^2}+4\!\times\!10^{-6}\sum_i r_i,\nonumber\\
\hat\theta&=\max(|\phi|,|\theta|)+0.006\,r_{sup}/1000.\nonumber
\end{align}
Here $r_{opp}$ is the opposite-motor residual, $r_{sup}=\sum_i r_i-r_{opp}$, and $p_m$ is the implemented motor-specific overboost term. Candidate scores combine threshold exceedance in these surrogates with effort, reference-pattern, support, saturation, and motor-specific penalties. The baseline thresholds are 0.330/0.350~m/s with $(w_v,w_h,w_d,w_\theta)=(1800,5000,80,0.08)$ and $(w_e,w_r,w_s,w_{sat})=(2\!\times\!10^{-9},10^{-7},2\!\times\!10^{-8},20)$. QP-lite is therefore a finite empirical candidate search with a quadratic-style score; it is not a continuous CBF-QP and no formal safety or global-optimality claim is made.

CEM \cite{rubinstein2004cem} tunes 13 score parameters offline over all four motors and $\eta\in\{0.4960,0.4965,0.4970,0.4975,0.4980\}$ using randomized near-hover surrogate states. The tuning script includes an empirical boundary-bias term derived from prior CrazySim observations; this term is used only during offline score tuning. The selected JSON configuration is frozen before the final simulator evaluations. No CEM optimization, online adaptation, or learned direct motor policy runs during the reported trials.

\subsection{Frozen Supervisor}
The development comparison defines $g(1)=g(3)=g(4)=\pi_{\mathrm{CEM}}$ and $g(2)=\pi_{\mathrm{PINV}}$. This map is not re-estimated on the fresh holdout, integrated-supervisor evaluation, or plant-sensitivity experiments.

%% file: sections/05_experiments.tex
\section{Experimental Methodology}
\label{sec:experiments}

\subsection{Environment and Protocol}
Experiments use CrazySim/Crazyflie SITL with Gazebo Sim 7.9, ROS~2 Humble, Ubuntu~22.04, and Python~3.10. Repeated trials use the deterministic seeded protocol \texttt{seeded\_ic\_v1}: $x_0,y_0\sim\mathcal{U}(-0.03,0.03)$~m; hover height is seeded within $0.70\pm0.02$~m; and fault time within $10.0\pm0.25$~s. A yaw perturbation is generated and logged by the seed utility but is not applied by the current launcher, so yaw is not counted as an applied randomized condition. Table~\ref{tab:experimental_protocol} summarizes the 3,570 executions in the primary evaluation set.

\begin{table*}[!t]
\centering\scriptsize
\caption{Primary evaluation protocol. Development trials are separated from frozen-policy validation and robustness experiments. The total excludes offline CEM surrogate evaluations, exploratory/pilot runs, and the nonretained 120-trial supervisor-v2 follow-up.}
\label{tab:experimental_protocol}
\begin{tabular}{l l r r}
\toprule
Experiment & Configuration & Replication & Trials\\
\midrule
Complementarity & 4 motors $\times$ 3 allocators, $\eta=.496$ & 30 matched/motor & 360\\
Fresh holdout & frozen routing & 30/motor & 120\\
Integrated supervisor & frozen routing, randomized order & 30/motor & 120\\
Nominal M2 boundary & 7 $\eta$ values & 30/$\eta$ & 210\\
OFAT sensitivity & 10 plants $\times$ 7 local $\eta$ & 30/$\eta$ & 2100\\
M2 routing stability & 11 plants $\times$ PINV/CEM & 30 paired & 660\\
\midrule
\multicolumn{3}{r}{\textbf{Total}} & \textbf{3570}\\
\bottomrule
\end{tabular}
\end{table*}

\subsection{Development, Validation, and Boundary Estimation}
At $\eta=0.496$, PINV, QP-lite, and CEM-tuned QP-lite are evaluated on 30 matched seeds per failed motor. These 360 executions select the routing rule and are therefore development data, not independent validation. After freezing the map, a fresh 120-trial holdout and a separate 120-trial randomized-order integrated-supervisor experiment use new seeds.

For M2/PINV, 30 trials are run at each $\eta\in\{0.49520,0.49530,\ldots,0.49580\}$. Firth bias-reduced logistic regression \cite{firth1993bias,heinze2002separation} fits
\begin{equation}
\operatorname{logit}P(\mathrm{safe}|\eta)=\beta_0+\beta_1(\eta-\eta_c)/10^{-4},
\end{equation}
so $\eta_{50}=\eta_c-10^{-4}\beta_0/\beta_1$. The analysis files record 5,000 paired-seed bootstrap replicates with seed 20260817. Lower $\eta_{50}$ denotes tolerance to more severe LoE.

\subsection{Plant Sensitivity and Routing Stability}
Mass, thrust coefficient, motor time constant, thrust-to-torque ratio, and arm length are perturbed independently by $\pm10\%$. Each of the ten OFAT plants uses a seven-point local $\eta$ grid and 30 seeds per point (2,100 trials). Sensitivity is $\Delta\eta_{50}^{(c)}=\eta_{50}^{(c)}-\eta_{50}^{(nom)}$. Unsafe-trial violations of vertical speed, horizontal speed, tilt, angular rate, and drift are recorded nonexclusively; these counts describe observed first-contact failure criteria and are not an optimization active-set analysis.

Finally, M2 PINV and frozen CEM are compared at $\eta=0.496$ under the nominal plant and all ten OFAT perturbations using 30 paired seeds per controller/state (660 executions). Exact McNemar tests summarize within-state paired outcomes. Aggregate validation rates use 95\% Wilson score intervals.

%% file: sections/06_results.tex
\section{Results}
\label{sec:results}

\subsection{Allocator Complementarity and Frozen Validation}
\begin{figure}[!t]
\centering
\includegraphics[width=\columnwidth]{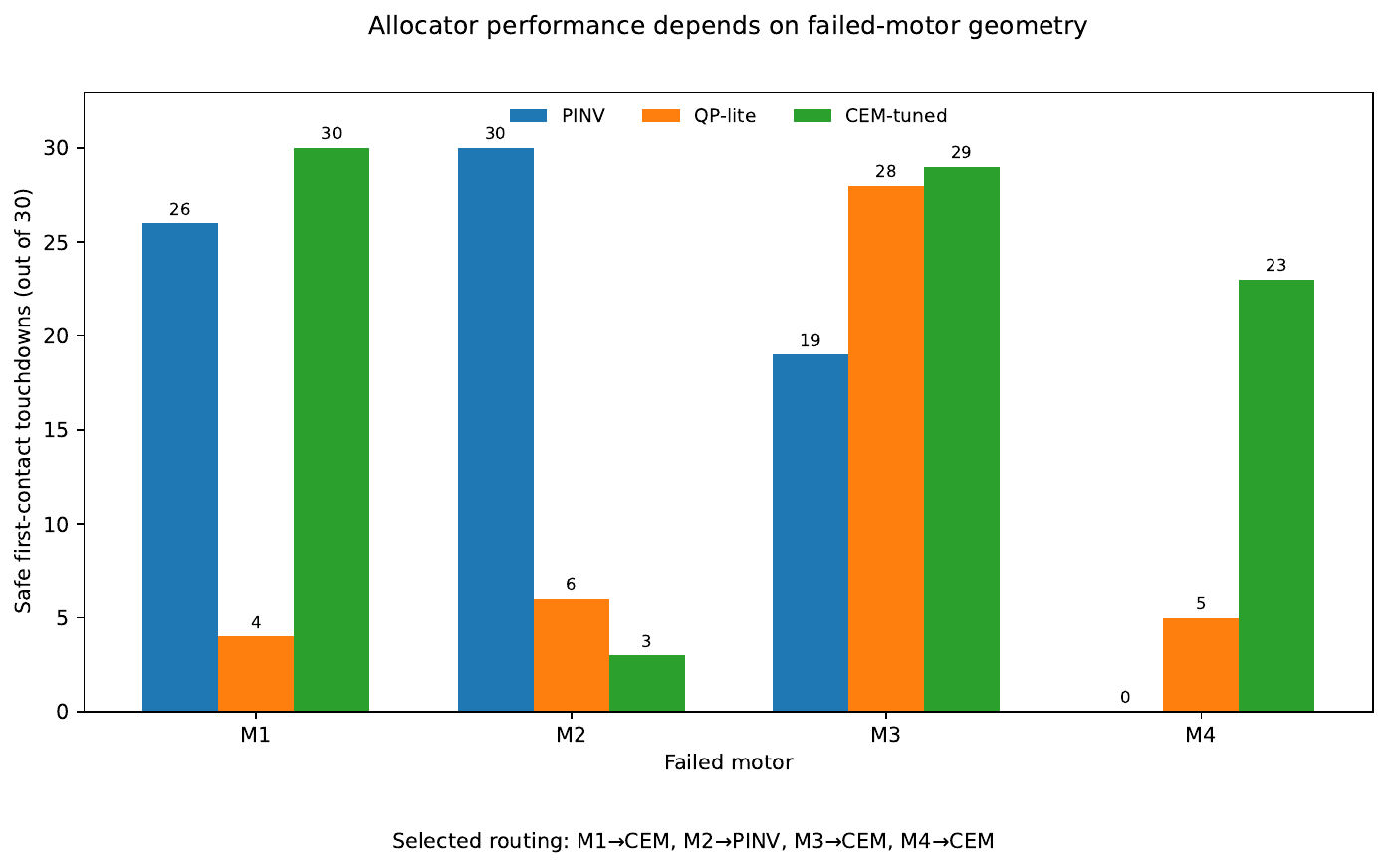}
\caption{Matched development comparison at $\eta=0.496$ (30 seeds per motor). The M3 CEM selection is 29/30 versus 28/30 for QP-lite.}
\label{fig:complementarity}
\end{figure}

At $\eta=0.496$, the development comparison depends strongly on failed-motor location (Fig.~\ref{fig:complementarity}). Safe counts for PINV/QP-lite/CEM are M1: 26/4/30, M2: 30/6/3, M3: 19/28/29, and M4: 0/5/23. Thus the tested allocators do not have a common winner across all four motor locations. M3 is the weak selection: CEM exceeds QP-lite by only one trial. The routing rule chosen from these development data totals 112/120 (93.3\%) and is not treated as validation performance.

\begin{figure}[!t]
\centering
\includegraphics[width=\columnwidth]{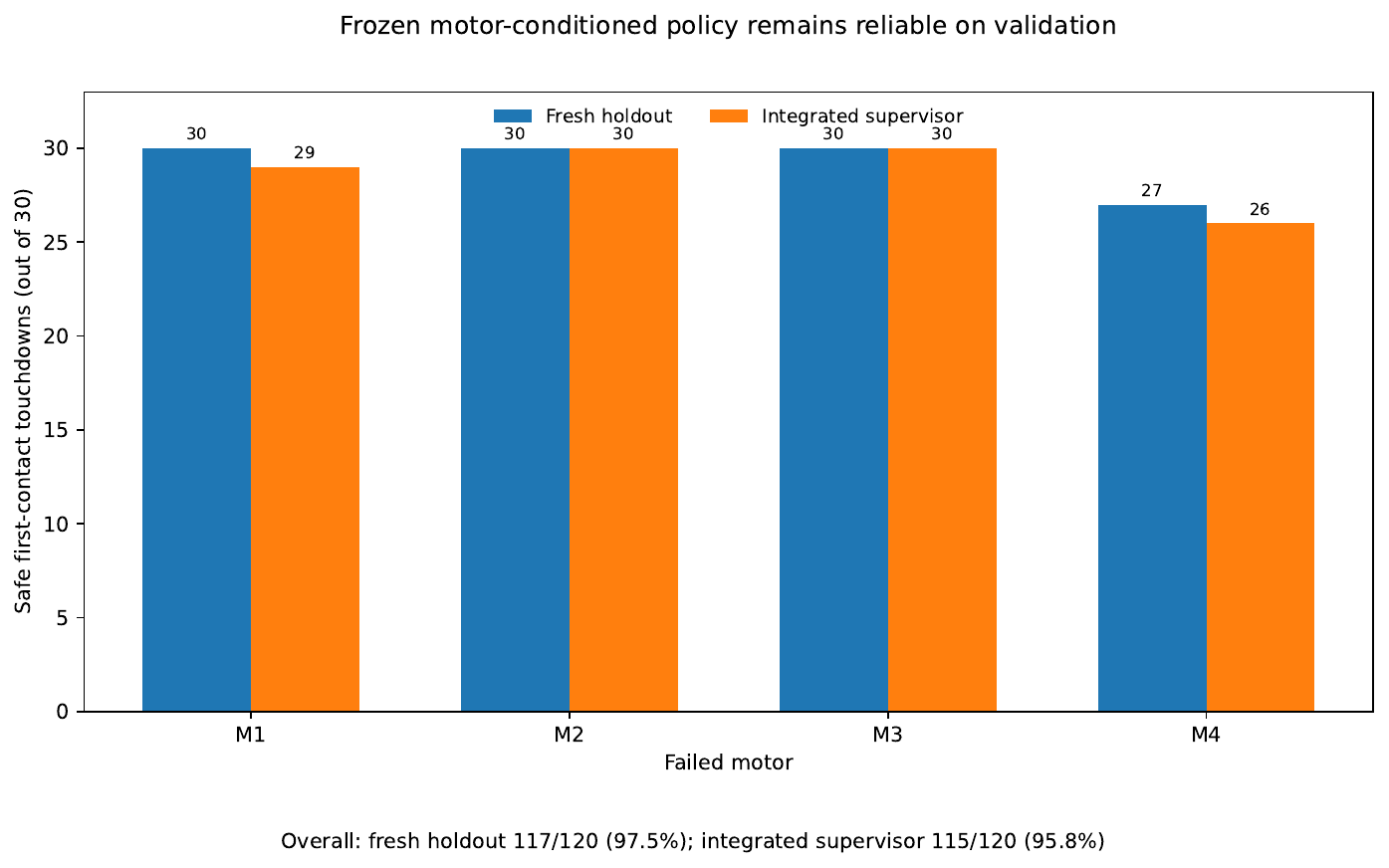}
\caption{Per-motor fresh-holdout and integrated-supervisor results for the frozen routing map. Failed-motor identity is supplied to the supervisor.}
\label{fig:validation}
\end{figure}

The frozen policy yields 117/120 (97.5\%, Wilson 95\% CI 92.9--99.1\%) on the fresh-seed holdout and 115/120 (95.8\%, 90.6--98.2\%) in the randomized-order integrated-supervisor evaluation (Fig.~\ref{fig:validation}). A separate, nonretained follow-up that replaced the M4 route by a manual opposite-M2 residual of 14,000 PWM yielded 113/120; it is not part of the reported frozen policy.

\subsection{Nominal M2/PINV Boundary}
\begin{figure}[!t]
\centering
\includegraphics[width=\columnwidth]{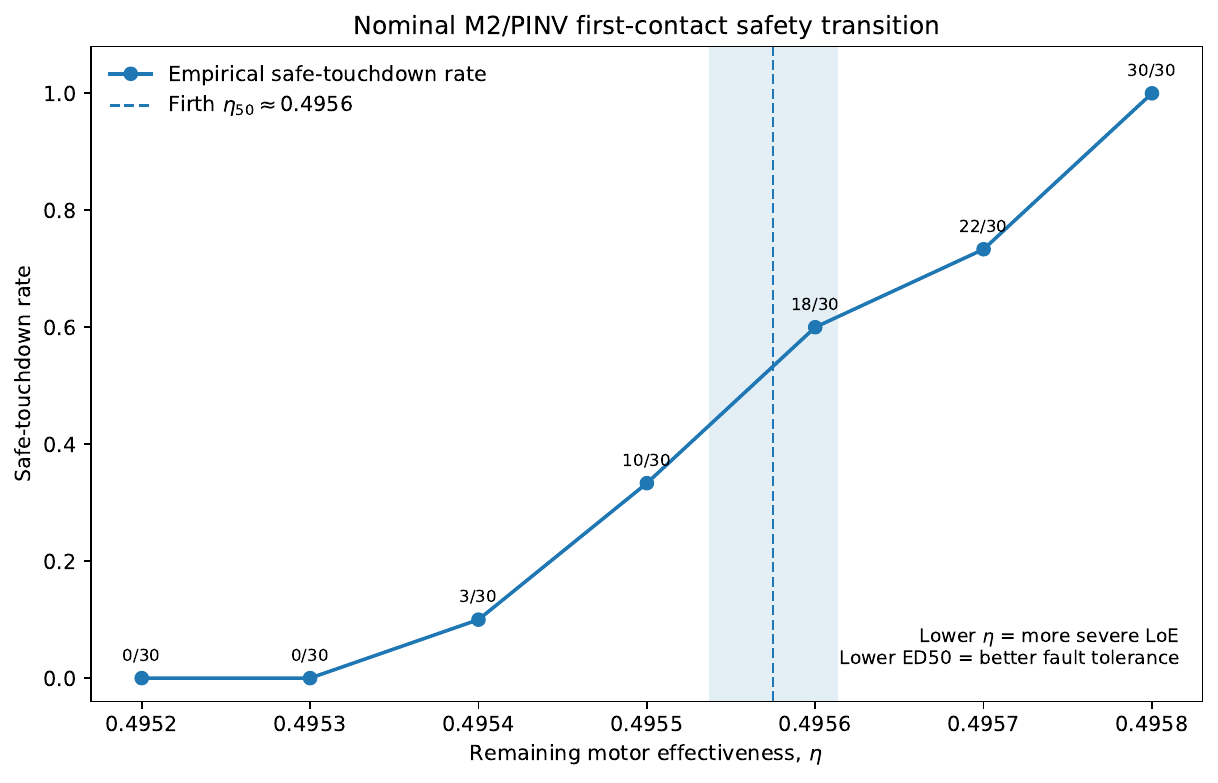}
\caption{Nominal M2/PINV first-contact safety transition. Because $\eta$ is remaining effectiveness, lower $\eta_{50}$ means tolerance to more severe LoE.}
\label{fig:boundary}
\end{figure}

The M2/PINV sweep gives empirical safe counts of 0, 0, 3, 10, 18, 22, and 30 out of 30 as $\eta$ increases from 0.49520 to 0.49580 (Fig.~\ref{fig:boundary}). Firth regression gives $\eta_{50}\approx0.4956$ with a within-model paired-seed bootstrap interval $[0.4955,0.4956]$. This interval quantifies repeated-seed uncertainty under the fixed nominal simulator model; it is not a confidence statement about physical plant parameters.

\subsection{Plant Sensitivity and Failure-Criterion Shift}
\begin{figure}[!t]
\centering
\includegraphics[width=\columnwidth]{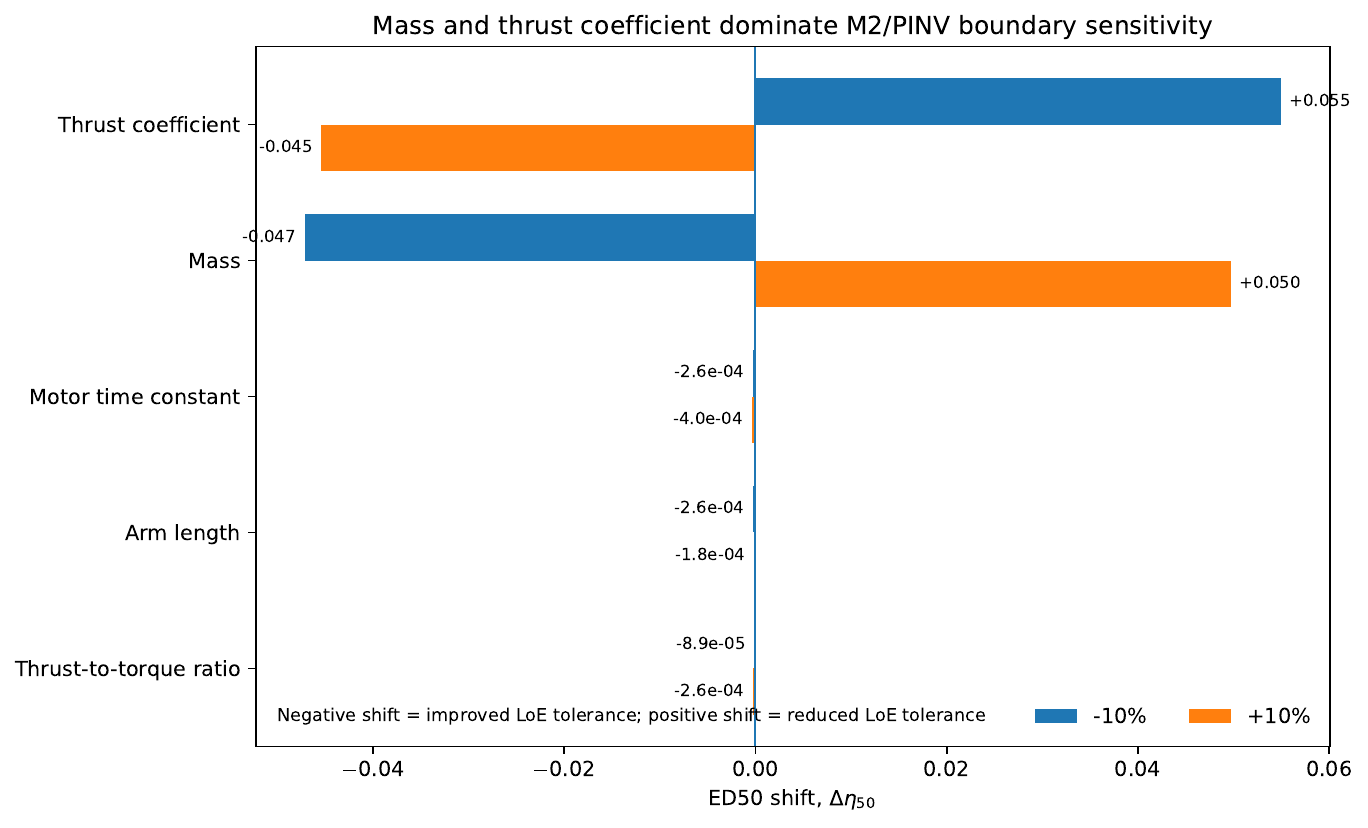}
\caption{M2/PINV $\eta_{50}$ shifts for the ten $\pm10\%$ OFAT plant perturbations.}
\label{fig:sensitivity}
\end{figure}

Thrust coefficient and mass produce the largest M2/PINV boundary displacements: their maximum $|\Delta\eta_{50}|$ values are 0.05495 and 0.04974, respectively, whereas the maximum for motor time constant, arm length, or thrust-to-torque ratio is 0.000404 (Fig.~\ref{fig:sensitivity}). Under the tested $\pm10\%$ OFAT perturbations, the mass/thrust shifts are therefore more than two orders of magnitude larger than the largest shift among the other three parameters.

\begin{figure}[!t]
\centering
\includegraphics[width=\columnwidth]{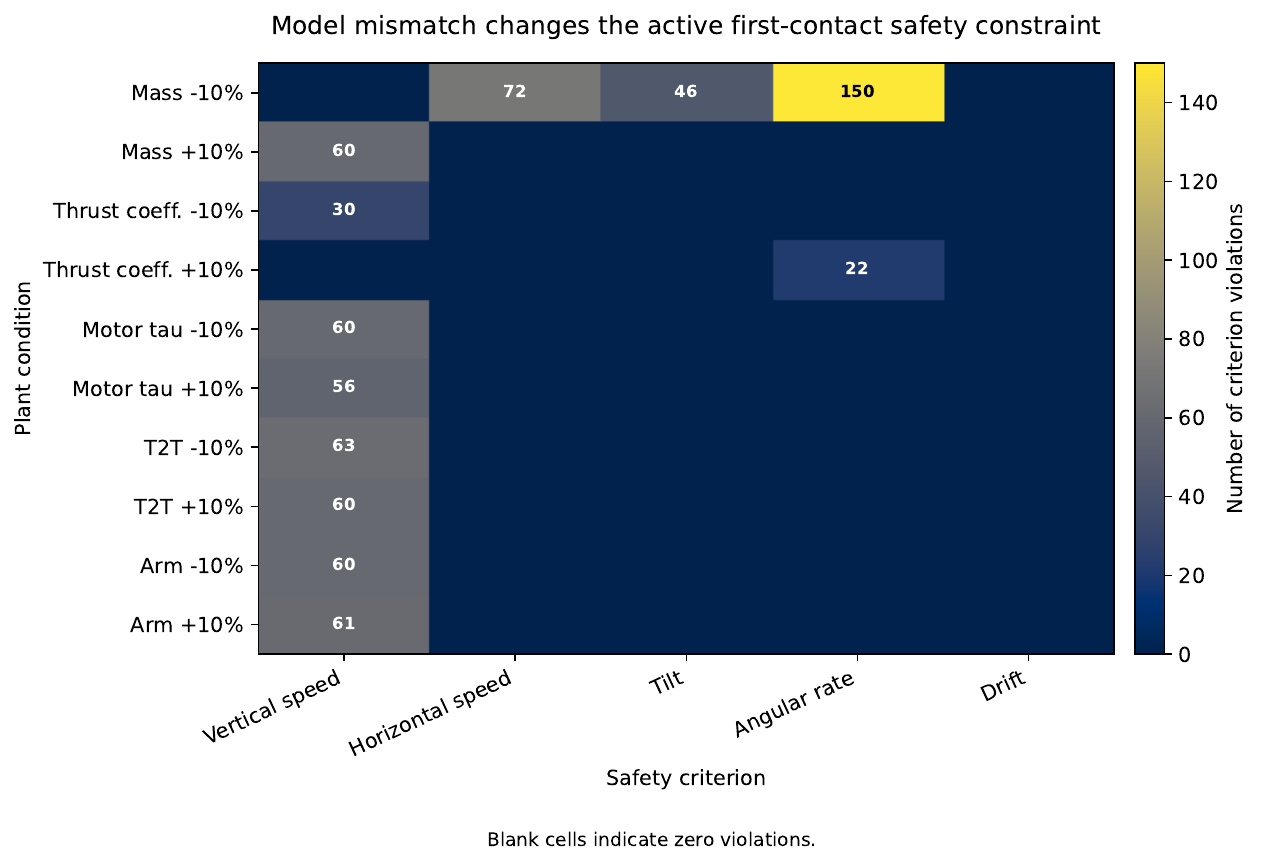}
\caption{Nonexclusive first-contact criterion-violation counts in the ten OFAT fine sweeps. Unannotated cells are zero.}
\label{fig:failure_heatmap}
\end{figure}

The observed violation pattern also changes (Fig.~\ref{fig:failure_heatmap}). For mass $-10\%$, the 210-trial fine sweep contains 150 angular-rate, 72 horizontal-speed, 46 tilt, and zero vertical-speed violations; counts overlap across criteria. Thrust coefficient $+10\%$ contains 22 angular-rate violations and zero violations of the other four recorded criteria. In the remaining eight OFAT conditions, vertical speed is the most frequent violated criterion.

\subsection{Routing Stability at Fixed $\eta$}
\begin{figure}[!t]
\centering
\includegraphics[width=\columnwidth]{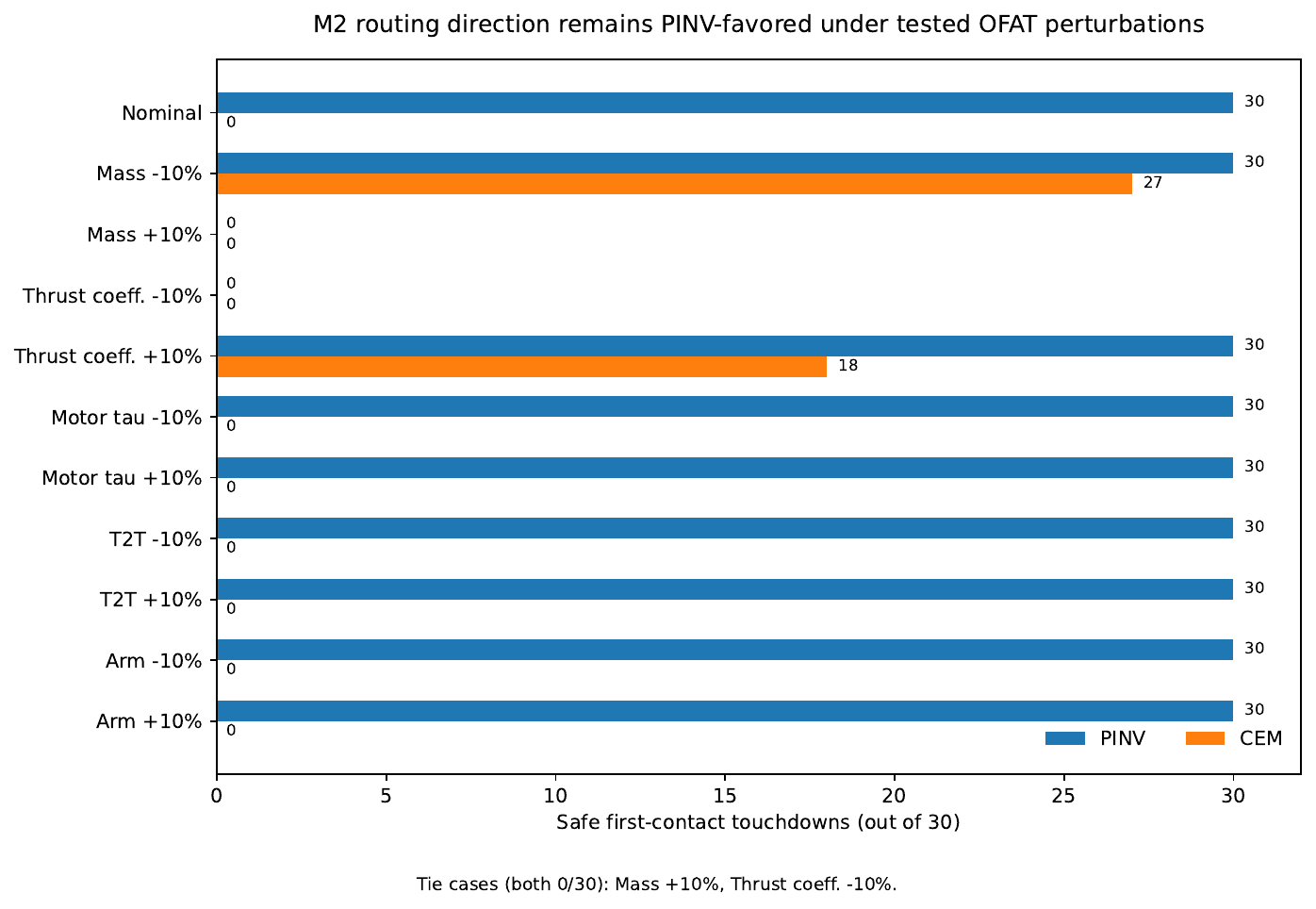}
\caption{M2 safe-touchdown counts for PINV and frozen CEM at $\eta=0.496$. The two 0/30--0/30 ties are shown explicitly.}
\label{fig:routing}
\end{figure}

At $\eta=0.496$, PINV has the higher M2 safe count in 9/11 tested plant states, ties CEM in 2/11, and is never lower (Fig.~\ref{fig:routing}). For mass $-10\%$, the paired counts are 30/30 versus 27/30 ($p=0.25$); for thrust coefficient $+10\%$, they are 30/30 versus 18/30 ($p=4.88\times10^{-4}$). The two ties are mass $+10\%$ and thrust coefficient $-10\%$, where both controllers are 0/30. Thus none of the ten tested perturbations reverses the M2 count ordering in favor of CEM, but two perturbations make both tested controllers fail all 30 first-contact safety trials.

\FloatBarrier

%% file: sections/07_discussion.tex
\section{Discussion and Limitations}
\label{sec:discussion}
The main experimental result is descriptive allocator complementarity under the tested single-motor LoE condition. M1 and M4 favor the CEM-tuned residual allocator by large development-set count margins, M2 favors bounded WLS/PINV, and M3 is nearly tied between the two residual variants. The frozen holdout and integrated-supervisor evaluations test that routing rule on new seeds within the same simulator family and fault specification; they do not establish transfer to hardware or to unknown fault identities. Likewise, the M2 OFAT experiment shows that the count ordering does not reverse in the ten tested $\pm10\%$ single-parameter perturbations, but it does not prove global routing optimality or a formal controllability boundary.

The study is simulation-only and assumes supplied failed-motor identity and effectiveness. Battery dynamics, hardware asymmetry, estimator error, contact-model mismatch, complete or simultaneous rotor failures, and online FDI are not evaluated. OFAT addresses one parameter at a time and does not characterize joint uncertainty. The bounded WLS allocator operates in PWM command-effectiveness coordinates rather than a calibrated physical wrench model. QP-lite searches a finite empirical residual library, and CEM tunes only its scoring parameters offline; neither method carries a formal safety or global-optimality guarantee. The first-contact violation counts identify which measured criteria fail most often, not which mathematical constraint is active in an optimization problem.

%% file: sections/08_conclusion.tex
\section{Conclusion}
Under the tested partial single-motor LoE condition, allocator ranking depends on failed-motor location: the development data select CEM-tuned residual allocation for M1/M3/M4 and bounded WLS/PINV for M2, with M3 separated from QP-lite by only one safe trial. After freezing the map, the policy obtains 117/120 safe first contacts on fresh seeds and 115/120 in the integrated supervisor. For M2/PINV, the nominal transition is sharply localized within the fixed simulator, while $\pm10\%$ mass and thrust-coefficient perturbations shift $\eta_{50}$ by about $5\times10^{-2}$. The most frequent violated first-contact criterion changes to angular rate for two OFAT conditions. At $\eta=0.496$, no tested plant perturbation gives CEM a higher M2 safe count than PINV, although two perturbations yield 0/30 for both. These results support motor-conditioned allocator selection within the evaluated simulation setting and identify plant-parameter sensitivity as a separate validation requirement.